\documentclass[letterpaper, 10 pt, conference]{ieeeconf}
\IEEEoverridecommandlockouts    
\usepackage{cite}
\usepackage{amsmath,amssymb,mathrsfs,amsfonts}
\usepackage{bbm}
\usepackage{algorithmic}
\usepackage{graphicx}
\usepackage{color}

\newcommand{\ad}{\mathrm{ad}}

\newcommand{\ve}{^\vee}

\begin{document}

\title{\LARGE \bf Estimating Infinite-Dimensional Continuum Robot States From the Tip}
\author{Tongjia Zheng$^1$, Ciera McFarland$^2$, Margaret Coad$^2$, and Hai Lin$^1$
\thanks{*This work was supported by the National Science Foundation under Grant No. CNS-1830335 and IIS-2007949.}
\thanks{$^1$Department of Electrical Engineering, University of Notre Dame, Notre Dame, IN 46556, USA (e-mail: tzheng1@nd.edu, hlin1@nd.edu). }
\thanks{$^2$Department of Aerospace and Mechanical Engineering, University of Notre Dame, Notre Dame, IN 46556, USA (e-mail: cmcfarl2@nd.edu, mcoad@nd.edu). }
}

\maketitle

\thispagestyle{empty}
\pagestyle{empty}

\begin{abstract}
Knowing the state of a robot is critical for many problems, such as feedback control. For continuum robots, state estimation is an incredible challenge. First, the motion of a continuum robot involves many kinematic states, including poses, strains, and velocities. Second, all these states are \textit{infinite-dimensional} due to the robot's flexible property. It has remained unclear whether these infinite-dimensional states are observable at all using existing sensing techniques. Recently, we presented a solution to this challenge. It was a mechanics-based dynamic state estimation algorithm, called a \textit{Cosserat theoretic boundary observer}, which could recover all the infinite-dimensional robot states by only measuring the velocity twist of the tip. In this work, we generalize the algorithm to incorporate tip pose measurements for more tuning freedom. We also validate this algorithm offline using experimental data recorded from a tendon-driven continuum robot. We feed the recorded tendon force and tip measurements into a numerical solver of the Cosserat rod model based on our robot. It is observed that, even with purposely deviated initialization, the state estimates by our algorithm quickly converge to the recorded ground truth states and closely follow the robot's actual motion.
\end{abstract}


\section{Introduction}

Continuum robots are made of compliant materials and can deform continuously along their length \cite{webster2010design}.
Thanks to the flexible deformation capability, continuum robots exhibit great potential in many applications, such as medical surgeries \cite{burgner2015continuum} and manipulation in constrained environments \cite{walker2013continuous}. 

In many robot-related problems, such as motion planning and control, it is important for a robot to know its current state.
For continuum robots, state estimation is incredibly challenging. 
To understand the difficulty, we may conceptually think of a continuum robot as the limit of a rigid-link robot that consists of infinitely many infinitesimal rigid cross-sections stacked along a centerline.
By this analogy, the motion of a continuum robot inevitably involves many kinematic states, such as positions, orientations, linear velocities, and angular velocities. It also involves bending, torsion, shear, and elongation, which can be viewed as the ``joint variables'' of the continuum robot. All these state variables vary continuously along the robot's length. In this regard, we say that the robot states are \textit{infinite-dimensional}. This makes its state estimation problem especially challenging.

\begin{figure}[tb]
    \centering
    \includegraphics[width=1\columnwidth]{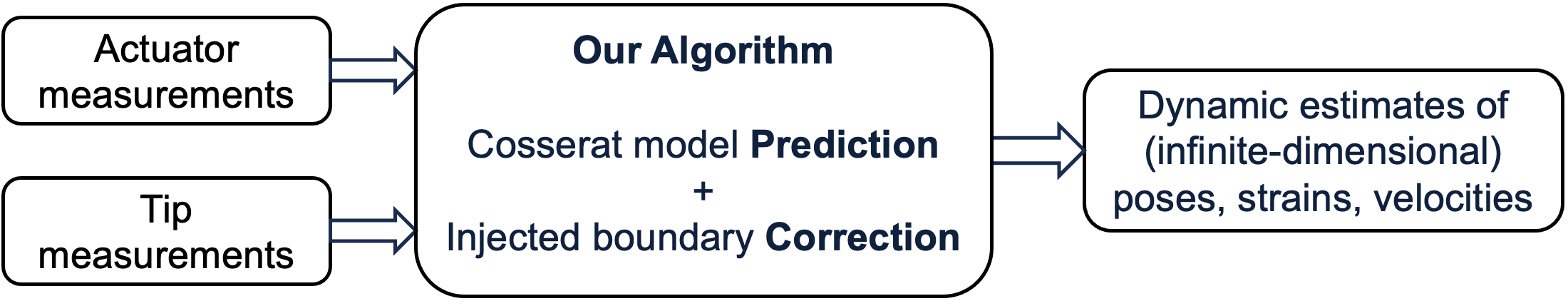}
    
    \includegraphics[width=1\columnwidth]{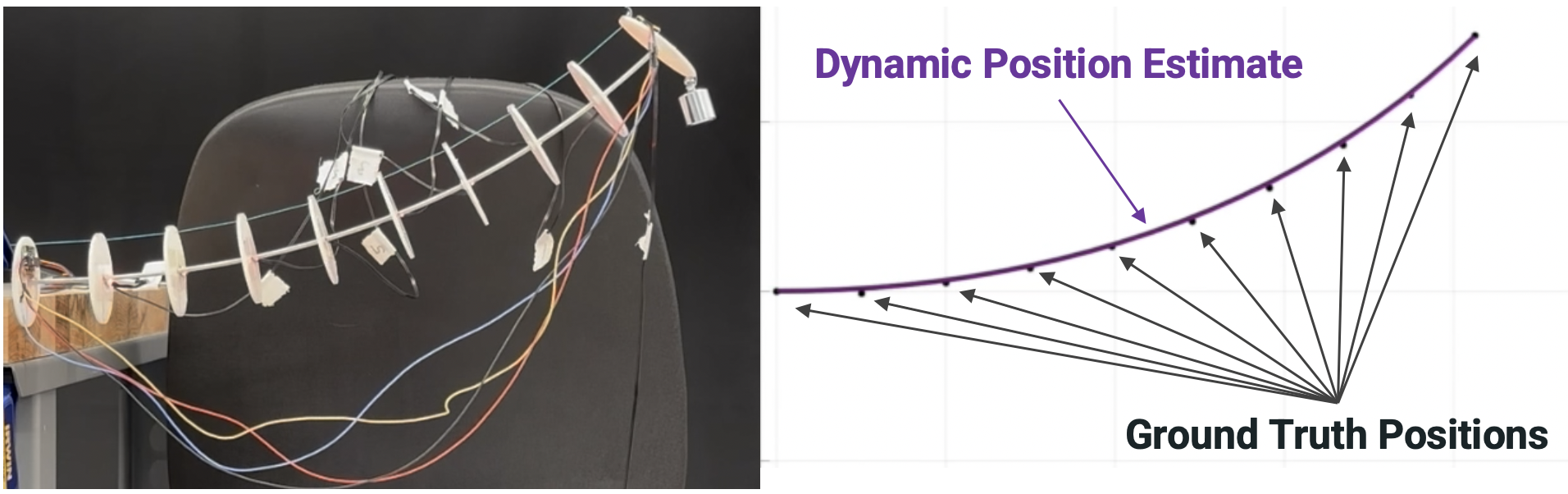}
    \caption{Overview of our Cosserat theoretic boundary observer algorithm and experimental results. This algorithm predicts the robot state based on the Cosserat rod model, actuator force measurements, and injected correction based on tip pose and velocity measurements. Experimental results show that the state estimates for the poses and velocities along the robot closely follow the ground truth states.}
    \label{fig:overview}
\end{figure}

An overview of the existing results can be found in \cite{lilge2022continuum}. 
We observe that existing approaches typically require a large amount of sensing, and even with that, it still seems difficult to conduct theoretical studies on the convergence of these algorithms.
A more fundamental question is whether these infinite-dimensional robot states are \textit{observable} at all based on existing sensing techniques.
A potential solution to this challenge was reported in our previous work \cite{zheng2023full}, where we designed a dynamic state estimation algorithm and showed that it could recover all these infinite-dimensional states by only measuring the velocity twist of the tip.
The algorithm was called a \textit{Cosserat theoretic boundary observer} because it was based on the Cosserat rod theory \cite{rucker2011statics,renda2014dynamic} and only required boundary (tip) sensing.

In this work, we generalize the algorithm in \cite{zheng2023full} to further incorporate tip pose measurements.
This provides more tuning freedom for the algorithm's performance.
We also validate our algorithm offline using recorded experimental data of a tendon-driven continuum robot.
We record the position trajectories of 10 markers on the backbone, which are used as ground truth. We also record the tendon's tension and the tip poses and velocities, which are then fed to our algorithm. We use SoRoSim \cite{mathew2022sorosim}, a numerical solver of the Cosserat rod model, to implement our algorithm. We observe that, even with purposely deviated initialization, the state estimates by our algorithm are still able to quickly converge to the recorded ground truth states and exhibit close tracking of the robot's actual motion. An overview of our algorithm and experimental results is given in Figure~\ref{fig:overview}.


\section{Background}

Currently, robot state estimation problems are typically solved using model prediction, sensing, or their combination.

If one has a precise dynamic model for the continuum robot and can exactly measure its initial state and the wrench inputs applied by actuators, then one can iterate the model to predict its state trajectories \cite{till2019real}.
This is known as \textit{model prediction}.
However, these assumptions are demanding due to the flexible property of the robot.
Thus, one typically observes accumulated deviations from the robot's actual motion over time if only using model prediction.

A complementary strategy is to use sensing.
An existing approach is to take many discrete measurements along the continuum robot using, for example, motion capture systems, IMUs, strain sensors, etc., and then fit a geometric model (such as a parametrized curve) to these discrete measurements \cite{song2015electromagnetic, bezawada2022shape}.
This approach usually only focuses on the primary strains (curvature) and neglects the others due to the insufficient amount of measurements.
Combining a mechanical model usually significantly improves the estimation performance, as it provides an inherent relation between the measurements.
One of the most widely accepted mechanical models for continuum robots is based on the Cosserat rod theory, which describes the dynamics of continuum robots using partial differential equations (PDEs).
Since state estimation of PDEs is relatively uncharted territory, existing work has to perform model reduction to convert it into a state estimation problem of systems of ordinary differential equations (ODEs).
Recall that the states of a continuum robot are usually functions of time and the arc parameter along the centerline.
One reduction technique is to fix the time instant and view the arc parameter as an analogy of the time parameter in traditional robotic systems of ODEs \cite{anderson2017continuum,lilge2022continuum}.
In this way, many estimation algorithms for ODEs (where the measurements are taken along time) can be applied to the continuum robot at a fixed time instant (where the measurements are taken along the arc length).
However, this approach needs to assume the continuum robot is under slow-speed motions in order to ignore the impact of velocities. 
As a result, the aforementioned approaches are classified as \textit{(static) shape estimation}.
To account for velocities and perform \textit{(dynamic) state estimation}, an existing reduction technique is to discretize the Cosserat PDE model into a Lagrangian ODE model \cite{boyer2020dynamics} and then design an observer \cite{rucker2022task} or an extended Kalman filter \cite{loo2019h, stewart2022state}.

Compared to existing approaches, the main novelty of our algorithm is that it only requires tip sensing, and the convergence of estimation errors can be formally proven \cite{zheng2023full}.

\section{Cosserat Rod Model For Continuum Robots}
\label{section:modeling}

In this section, we introduce the Cosserat rod model for continuum robots \cite{rucker2011statics, renda2014dynamic}.
To avoid the massive use of cross products in the equations, we will make use of some Lie group notation in $SO(3)$ (the group of rotation matrices) and $SE(3)$ (the group of homogeneous transform matrices) \cite{murray2017mathematical}.
The definitions are included in the Appendix.

\begin{figure}[tb]
    \centering
    \includegraphics[width=0.9\columnwidth]{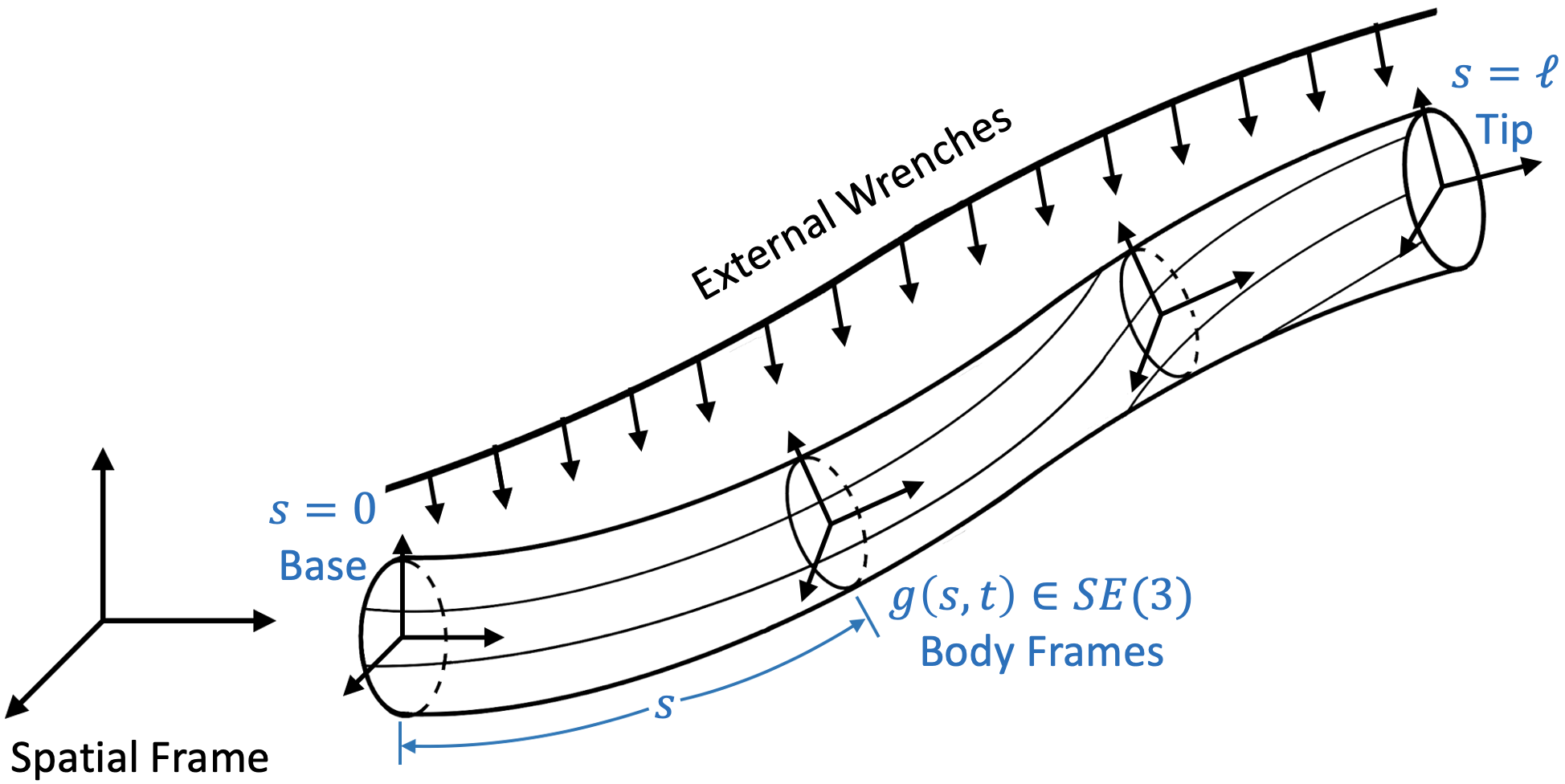}
    \caption{Our continuum robot model. Using Cosserat rod theory, our robot is modeled as a continuous set of rigid cross-sections stacked along a centerline parametrized by $s$ from the base ($s=0$) to the tip ($s=\ell$) where $\ell$ is the total length.
    $g(s,t)\in SE(3)$ is the pose (homogeneous transform matrix) of the cross-section at location $s$ and time $t$.
    Each cross-section defines a body frame.
    The robot undergoes internal wrenches (due to elastic deformation and actuation) and external wrenches (such as gravity).}
    \label{fig:Cosserat rod}
\end{figure}

The Cosserat rod theory models a continuum robot as a continuous set of rigid cross-sections stacked along a centerline; see Figure~\ref{fig:Cosserat rod}.
All states of the continuum robot are continuous functions of the arc parameter $s\in[0,\ell]$ and time $t$.
Let $p(s,t)\in\mathbb{R}^3$ be the position vector of the centerline and $R(s,t)\in SO(3)$ be the rotation matrix of each cross-section.
Note that each cross-section defines a body frame.
The pose of the robot is denoted by
\begin{align*}
    g(s,t)=
    \begin{bmatrix}
        R(s,t) & p(s,t) \\
        0_{1\times3} & 1
    \end{bmatrix}\in SE(3),
\end{align*}

To define kinematics, let $w(s,t),v(s,t),u(s,t),q(s,t)\in\mathbb{R}^3$ be the fields of angular velocities, linear velocities, angular strains, and linear strains, respectively, of the cross-sections in their body frames.
Let
\begin{align*}
    \eta(s,t)=
    \begin{bmatrix}
        w(s,t) \\
        v(s,t)
    \end{bmatrix},\quad
    \xi(s,t)=
    \begin{bmatrix}
        u(s,t) \\
        q(s,t)
    \end{bmatrix}
\end{align*}
be the fields of velocity twists and strain twists, respectively.
We abbreviate the partial derivatives according to $\partial_t=\frac{\partial}{\partial t}$ and $\partial_s=\frac{\partial}{\partial s}$.
We also omit the dependence of the variables on $s$ and $t$ when writing equations.
Then, the kinematics of the continuum robot are given by
\begin{align}
    \partial_tg=g\eta^\wedge, \label{eq:kinematics 1} \\
    \partial_sg=g\xi^\wedge. \label{eq:kinematics 2}
\end{align}
The equality of mixed partial derivatives $\partial_{st}g=\partial_{ts}g$ yields the compatibility equation
\begin{align}
    \partial_t\xi=\partial_s\eta+\ad_{\xi}\eta. \label{eq:compatibility}
\end{align}

To define dynamics, let $m(s,t),n(s,t),l(s,t),f(s,t)\in\mathbb{R}^3$ be the fields of internal moments, internal forces, external moments, and external forces, respectively, of the cross-sections in their body frames.
Let
\begin{align*}
    \Phi(s,t)=
    \begin{bmatrix}
        m(s,t) \\
        n(s,t)
    \end{bmatrix},\quad
    \Psi(s,t)=
    \begin{bmatrix}
        l(s,t) \\
        f(s,t)
    \end{bmatrix}
\end{align*}
be the fields of internal and external wrenches, respectively.
Applying Hamilton's principle in the context of Lie groups yields the following dynamics (expressed in the body frames) for the continuum robot
\begin{align} \label{eq:dynamics}
    J\partial_t\eta-\ad_\eta^TJ\eta=\partial_s\Phi-\ad_\xi^T\Phi+\Psi,
\end{align}
where $J(s)\in\mathbb{R}^{6\times6}$ is the cross-sectional inertia matrix, which can be a function of $s$ in the case of, e.g., nonuniform cross-sectional areas.
We have the following boundary conditions at $s=0$ (the base) and $s=\ell$ (the tip):
\begin{align}
    & \eta(0,t)=\eta_{\text{base}}(t), \quad \Phi(\ell,t)=\Phi_{\text{tip}}(t),
\end{align}
where $\eta_{\text{base}}(t)$ is the velocity twist of the base and $\Phi_{\text{tip}}(t)$ is the point wrench applied at the tip.

Fluidic and tendon actuation are usually modeled as distributed internal wrenches \cite{renda2020geometric}.
Thus, we have
\begin{align}
    \Phi(s,t) & =\phi(s,t)-\phi_{\text{act}}(s,t), \label{eq:Phi}
\end{align}
where $\phi_{\text{act}}(s,t)$ is the distributed internal wrench input applied by fluidic or tendon actuation and $\phi(s,t)$ is the field of wrenches due to elastic deformation.
The wrench $\phi(s,t)$ is usually assumed to satisfy a linear constitutive law:
\begin{align} \label{eq:linear constitutive law}
    \phi=K(\xi-\xi_*),
\end{align}
where $\xi_*(s)\in\mathbb{R}^6$ is the strain field of the undeformed reference configuration and $K(s)\in\mathbb{R}^{6\times6}$ is the cross-sectional stiffness matrix, which can be a function of $s$ as well.
There are more general constitutive laws that can involve material damping and nonlinear stress-strain relations.

\section{State Estimation of Continuum Robots}
\label{section:algorithm}

This section describes the setup of the estimation problem and our solution, the \textit{Cosserat theoretic boundary observer}.

\subsection{Problem Setup}

To define a state estimation problem, it is essential to understand the system states and identify the smallest possible set of independent states. This set of states can then be used to determine the remaining states.
For a Cosserat rod model, the system states include the poses $g(s,t)$, the strains $\xi(s,t)$, the velocities $\eta(s,t)$, and the elastic wrenches $\phi(s,t)$.
The minimum representation is given by either $\{g(s,t),\eta(s,t)\}$ or $\{\xi(s,t),\eta(s,t)\}$, because at every time instant, $g(s,t)$ and $\xi(s,t)$ can be uniquely determined from each other according to \eqref{eq:kinematics 2} and $\phi(s,t)$ can be uniquely determined by $\xi(s,t)$ according to \eqref{eq:linear constitutive law}.
For state estimation of continuum robots, we find it more convenient to estimate $\{\xi(s,t),\eta(s,t)\}$ first and use them to determine other intermediate states.

We assume robot parameters $J(s)$ and $K(s)$ are known.
The calculation of $J(s)$ and $K(s)$ will be illustrated in Section \ref{section:experiment}.
For simplicity, we assume the base of the robot is fixed, the tip is loaded with a known force $F_{\text{tip}}(t)\in\mathbb{R}^3$ (represented in the spatial frame), and the robot is actuated by tendons with known mounting patterns and known tensions measured using mounted force sensors.
In this way, the input wrench $\phi_{\text{act}}(s,t)$ applied by the tendons can also be calculated according to \cite{renda2020geometric}.
This simplification is mainly for illustrative purposes.
We refer to our previous work \cite{zheng2023full} for a comprehensive treatment of more general setups such as moving bases and other types of loading and actuation.

The assumptions so far are standard for robot state estimation problems.
What makes this work distinct is the assumption of external measurements.
The existing work typically requires a large number of external measurements, such as discrete position and/or rotation measurements along the robot to construct a reasonable estimate of the robot's configuration.
In this work, we only need to measure the pose $g_{\text{tip}}(t)\in SE(3)$ and velocity twist $\eta_{\text{tip}}(t)\in\mathbb{R}^6$ of the tip.
This is possibly the minimum amount of necessary sensing to reconstruct all the infinite-dimensional states (poses, strains, and velocities) of the robot; otherwise, the estimation problem may be ill-posed and a solution does not exist.
This type of sensing can be achieved by attaching a single marker and an IMU at the tip and using a motion capture system.

The state estimation problem is stated as follows.
Assuming we know the system parameters $\{J(s),K(s)\}$, the inputs $\{\phi_{\text{act}}(s,t),F_{\text{tip}}(t)\}$, and the tip measurements $\{g_{\text{tip}}(t),\eta_{\text{tip}}(t)\}$, we want to estimate $\{\xi(s,t),\eta(s,t)\}$ (and therefore $g(s,t)$).
Note that we do not assume the initial states are known, as they are usually not.

\subsection{Design of the Cosserat Theoretic Boundary Observer}

The spirit of the \textit{Cosserat theoretic boundary observer} is similar to that of the Kalman filter in that it recursively uses the Cosserat rod model and sequential inputs to predict the next state and at the same time, injects sequential tip measurements into the algorithm to steer the prediction toward the true states.
That is why it is called an ``observer.''
The term ``boundary'' accounts for the fact that external measurements are only taken at the boundary (the tip).

A hat $\wedge$ on top of a variable indicates that it is an estimate.
For example, $\hat{\eta}(s,t)$ is the estimate of the velocity $\eta(s,t)$.
Note that $\hat{(\cdot)}$ and $(\cdot)^\wedge$ are different notations.
The former is a state estimate while the latter is the hat operator (defined in the Appendix) accounting for cross products.
Recall that $s=\ell$ is the tip.
For clarity, we denote $\hat{g}_{\text{tip}}(t)=\hat{g}(\ell,t)$ and $\hat{\eta}_{\text{tip}}(t)=\hat{\eta}(\ell,t)$ which represent the estimates of the tip pose and velocity, respectively, at time $t$.
Similarly, we may write
\begin{align*}
    \hat{g}_{\text{tip}}(t)=
    \begin{bmatrix}
        \hat{R}_{\text{tip}}(t) & \hat{p}_{\text{tip}}(t) \\
        0_{1\times3} & 1
    \end{bmatrix}.
\end{align*}

The complete \textit{Cosserat theoretic boundary observer} is then given by the following equations:
\begin{align}
    \partial_t\hat{\xi} & =\partial_s\hat{\eta}+\ad_{\hat{\xi}}\hat{\eta}, \label{eq:boundary observer 1} \\
    J\partial_t\hat{\eta}-\ad_{\hat{\eta}}^TJ\hat{\eta} & =\partial_s\hat{\Phi}-\ad_{\hat{\xi}}^T\hat{\Phi}+\Psi, \label{eq:boundary observer 2}
\end{align}
with the following boundary conditions at $s=0$ (the base) and $s=\ell$ (the tip):
\begin{align}
    \hat{\eta}(0,t) & =0, \label{eq:BC 1} \\
    \hat{\Phi}(\ell,t) & =-\Gamma_{\text{P}}\mathrm{Err}\big(\hat{g}_{\text{tip}}(t),g_{\text{tip}}(t)\big)-\Gamma_{\text{D}}\big(\hat{\eta}_{\text{tip}}(t)-\eta_{\text{tip}}(t)\big) \nonumber \\
    & \quad +\begin{bmatrix}
        0 \\
        \hat{R}_{\text{tip}}^T(t)F_{\text{tip}}(t)
    \end{bmatrix}, \label{eq:BC 2}
\end{align}
where the intermediate variables $\hat{\Phi}(s,t),\hat{g}(s,t)$ are computed from $\hat{\xi}(s,t)$ at every $t$ according to
\begin{align}
    \hat{\Phi} & =K(\hat{\xi}-\xi_*)-\phi_{\text{act}}, \label{eq:intermediate variable 1} \\
    \partial_s\hat{g} & =\hat{g}\hat{\xi}^\wedge, \label{eq:intermediate variable 2}
\end{align}
$\mathrm{Err}\big(\hat{g}_{\text{tip}}(t),g_{\text{tip}}(t)\big)$ is a function (defined later) that represents the difference between $\hat{g}_{\text{tip}}(t)$ and $g_{\text{tip}}(t)$, and $\Gamma_{\text{P}},\Gamma_{\text{D}}\in\mathbb{R}^{6\times6}$
are positive definite matrices representing the observer gains for adjusting the performance of the observer.

We provide a few useful insights about the boundary observer (Eqns. \ref{eq:boundary observer 1}-\ref{eq:intermediate variable 2}).
\begin{enumerate}
    \item It simply consists of two parts: prediction using the Cosserat rod model and correction through the boundary condition \eqref{eq:BC 2} using the tip measurements.
    \item If one thinks of it as a controller, then the correction term $-\Gamma_{\text{P}}\mathrm{Err}(\hat{g}_{\text{tip}},g_{\text{tip}})-\Gamma_{\text{D}}(\hat{\eta}_{\text{tip}}-\eta_{\text{tip}})$ is essentially a proportional-derivative (PD) controller.
    Compared to the D controller in our previous work \cite{zheng2023full}, the additional P term provides more flexibility to adjust performance.
    \item It can be initialized with any reasonable configuration, such as a straight or an equilibrium configuration.
    The correction term will consistently steer the estimation errors toward convergence over time.
    \item The physical intuition is that the tip estimation errors are injected into the algorithm as virtual tip damping that dissipates the energy of the estimation errors of the entire robot.
    This intuition has been used in \cite{zheng2023full} to prove the convergence of the observer in the D case.
    \item Since it is essentially still a Cosserat rod model with a virtual tip load, it can be numerically implemented using any existing solvers for the Cosserat rod model.
\end{enumerate}

Now we discuss how to define the pose estimation errors $\mathrm{Err}\big(\hat{g}_{\text{tip}},g_{\text{tip}}\big)$.
Recall that $\hat{g}_{\text{tip}},g_{\text{tip}}\in\mathbb{R}^{4\times4}$ are homogeneous transform matrices and we want $\mathrm{Err}\big(\hat{g}_{\text{tip}},g_{\text{tip}}\big)$ to be a column vector in $\mathbb{R}^6$.
This problem has been extensively studied under the name of ``geometric control'' \cite{bullo2019geometric}.
We simply list two popular choices among others.
One way is to define
\begin{align*}
    \mathrm{Err}\big(\hat{g}_{\text{tip}},g_{\text{tip}}\big)=
    \begin{bmatrix}
        \Big(\hat{R}_{\text{tip}}^TR_{\text{tip}}-R_{\text{tip}}^T\hat{R}_{\text{tip}}\Big)\ve \\
        \hat{p}_{\text{tip}}-p_{\text{tip}}
    \end{bmatrix},
\end{align*}
or alternatively,
\begin{align*}
    \mathrm{Err}\big(\hat{g}_{\text{tip}},g_{\text{tip}}\big)=\Big(\log\hat{g}_{\text{tip}}^{-1}g_{\text{tip}}\Big)\ve,
\end{align*}
where $\vee$ is the inverse operator of $\wedge$ defined in the Appendix.
Both definitions result in a column vector in $\mathbb{R}^6$ which represents the difference between $\hat{g}_{\text{tip}}$ and $g_{\text{tip}}$.

Finally, notice that we have simply assumed a linear constitutive law in \eqref{eq:intermediate variable 1}.
If a more accurate constitutive law is known, it can be substituted into \eqref{eq:intermediate variable 1}.
However, the current linear constitutive law works well in our experiments possibly because PD ``controllers'' have been empirically proven to be robust to model uncertainties.

\section{Experimental Design and Validation}
\label{section:experiment}

In this section, we introduce the robot's design and describe the experiments conducted to demonstrate the algorithm's ability to accurately recover the robot's motion.

\subsection{Robot Design and Experimental Setup}

The robot consisted of an aluminum 6061 rod (McMaster, Elmhurst, IL) with a radius of 1.6 mm acting as the backbone and ten disks glued every 50 mm that we 3D-printed using tough polylactic acid (tough PLA) on a fused deposition modeling 3D printer (S5, Ultimaker, Netherlands). 
The nominal properties of the rod are shown in Table \ref{tab:robot parameter}. 



\begin{table}[h]
\caption{Properties of our robot backbone.}
\small
\setlength{\tabcolsep}{1em}
\renewcommand{\arraystretch}{1.2}
    \centering
    \begin{tabular}{ll}
        \hline
        Length $\ell$ & $450~\mathrm{mm}$ \\
        Radius $r$ & $1.6~\mathrm{mm}$ \\
        Density $\rho$ & $2700~\mathrm{kg/m^3}$ \\
        Young's modulus $E$ & $68.9~\mathrm{GPa}$ \\
        Poisson's ratio $\nu$ & $0.325$ \\
        Shear modulus $G$ & $26~\mathrm{GPa}$ \\
        \hline
    \end{tabular}
    \label{tab:robot parameter}
\end{table}

\begin{figure}[t]
    \centering
    \includegraphics[width=0.5\columnwidth]{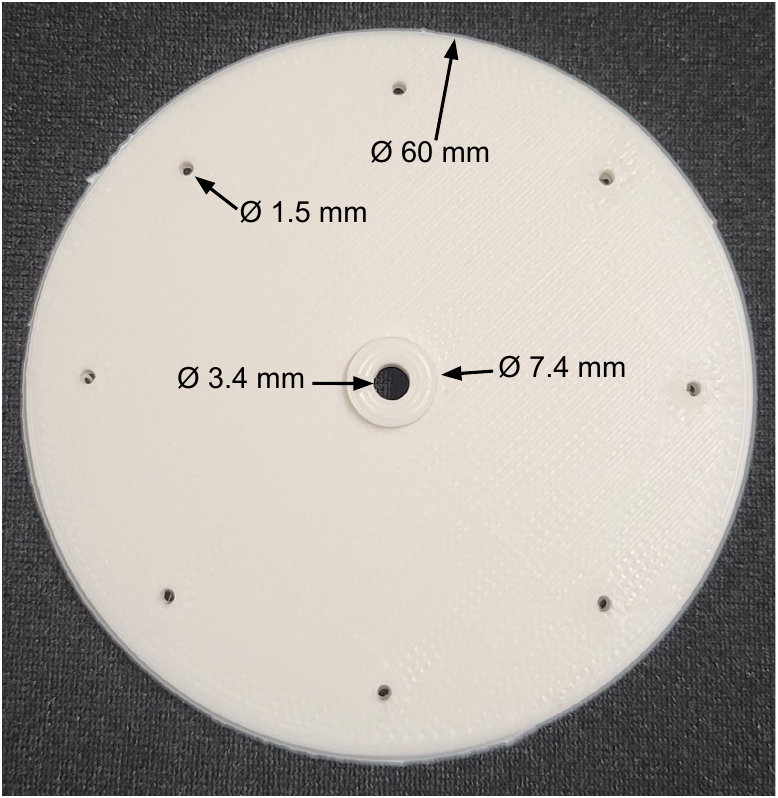}
    \caption{Planar view of a single disk, 10 of which are attached along our robot backbone. The ridge at the central hole extends 1 mm from the surface of the disk and is located on the side of the disk facing the robot base. The small holes are 25 mm from the center. Each disk's thickness is 2 mm.}
    \label{fig:Disk Photo}
\end{figure}

Each disk had a diameter of 60 mm and a thickness of 2 mm. The central hole in which the rod was placed had a diameter of 3.4 mm. We included a ridge around the central hole that was 7.4 mm in diameter and extruded 1 mm above the surface of the disk. 
We placed eight equally spaced holes on the disk that were 1.5 mm in diameter and 25 mm from the center of the disk. 
The tendon was made of a polyethylene fishing line with a diameter of 0.56 mm. 
For all experiments, the tendon was located in the top hole of each disk. 
The disk design can be viewed in Figure~\ref{fig:Disk Photo}. 

\begin{figure}[h]
    \centering
    \includegraphics[width=0.8\columnwidth]{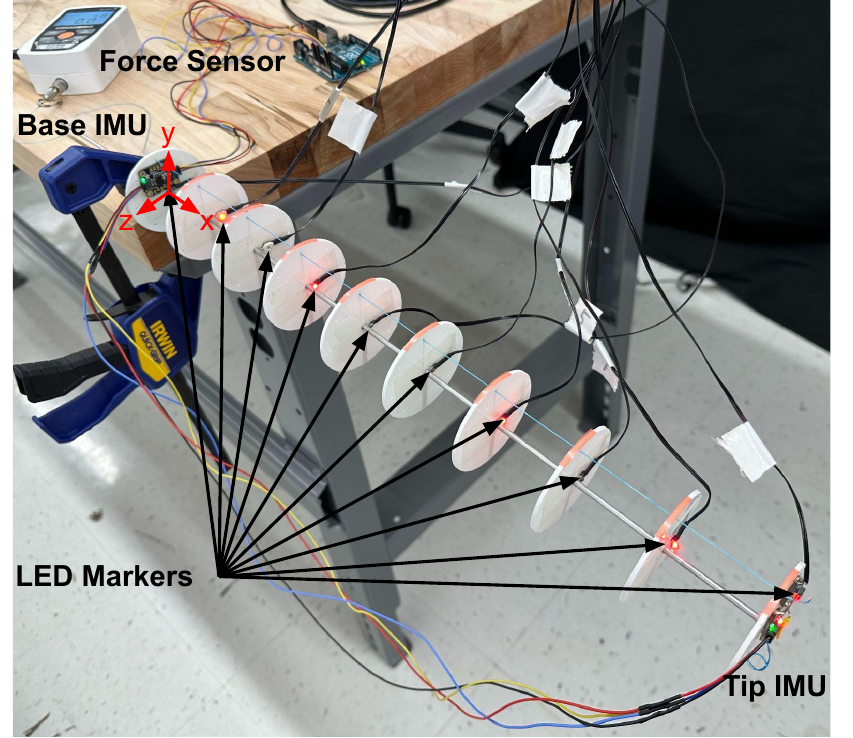}
    \caption{Setup for experiments. We placed one LED tracking marker at the center of each of the ten disks to measure their ground truth position. We placed one IMU on each of the base and tip disks to measure the tip orientation relative to a known reference orientation. We used a force sensor to measure the tension of a tendon routed through the holes on top of the 10 disks.
    The body frame of the base disk was defined by the red axes attached to the base disk, which was also chosen to be the spatial frame.
    The body frames attached to all other cross-sections were similarly defined.}
    \label{fig:Setup Diagram}
\end{figure}


To establish a ground truth model of the robot's position, we placed LED markers on each of the disks, which communicated with our motion capture system (Impulse X2E, PhaseSpace, San Leandro, CA). Additionally, we placed an IMU (BNO055, Adafruit, New York City, NY) on the tip of the robot to measure the robot's tip orientation. We used a second IMU placed at the robot base to establish a reference frame for the tip one. We tied a force sensor (M3-20, Mark-10, Copiague, NY) to the tendon to measure its tension. All components of the experimental setup are shown in Figure~\ref{fig:Setup Diagram}. 

\begin{figure*}[t]
    \centering
    \includegraphics[width=1\textwidth]{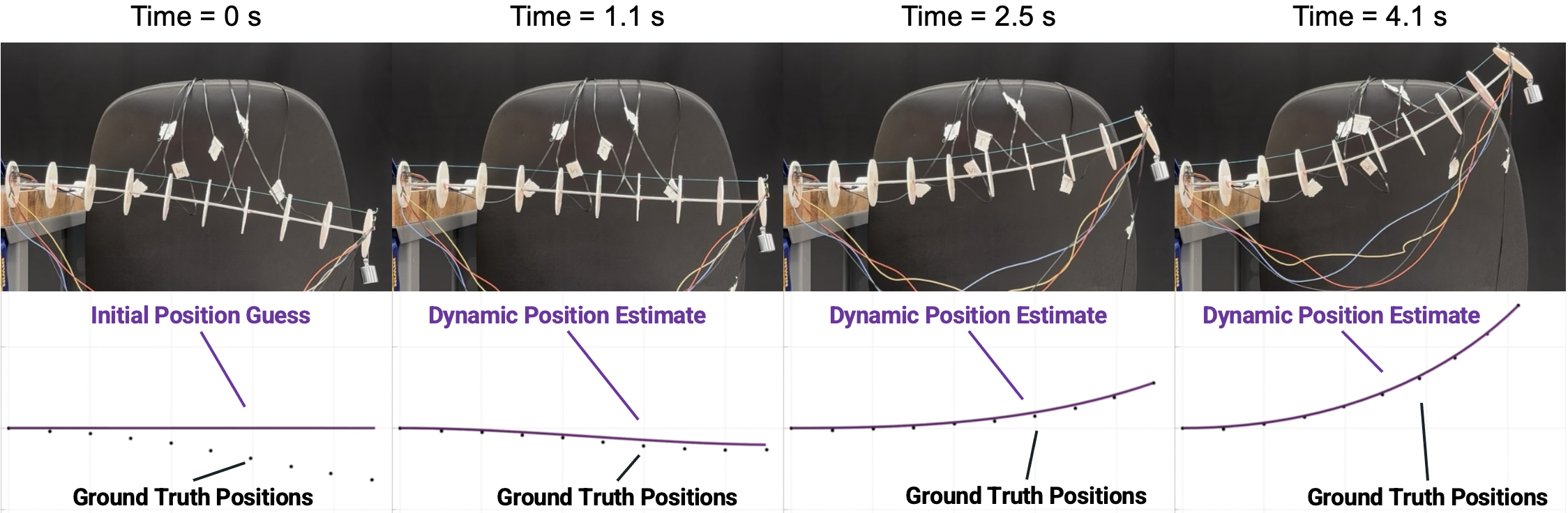}
    \caption{Snapshots of the experiment and offline validation results of the algorithm. In the top row, the robot exhibited planar motions under a 50~g tip load and the time-varying tendon tension. In the bottom row, the black dots are the ground truth positions of the markers on the backbone, and the purple curve is the estimated position of the backbone by our algorithm. We purposely initialized our algorithm with a deviated configuration. The estimated positions quickly converged to the ground truth positions and exhibited close tracking of the robot's actual motion.}
    \label{fig:estimation}
\end{figure*}

\begin{figure}[t]
    \centering
    \includegraphics[width=0.98\columnwidth]{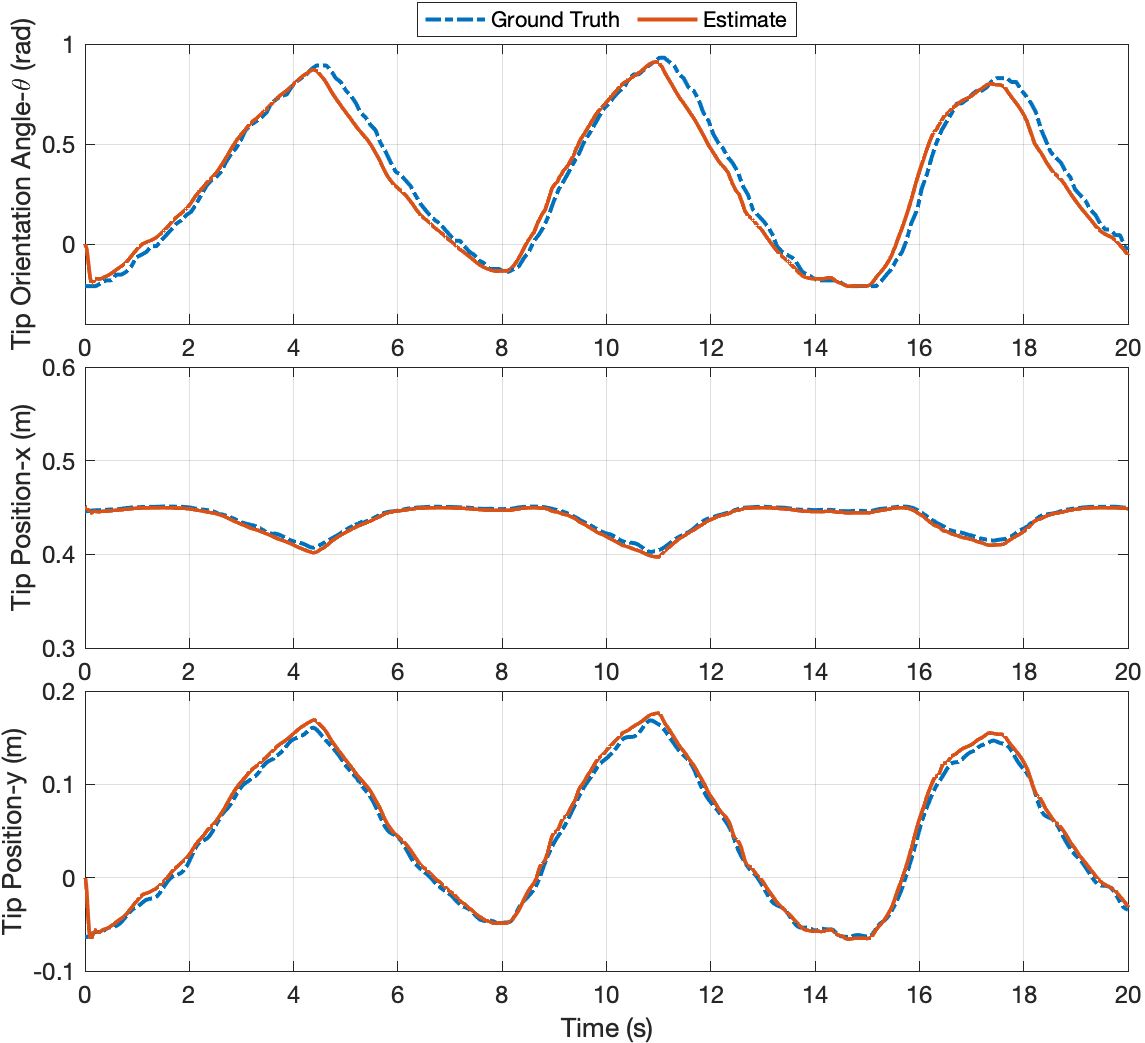}
    \caption{Comparison of the (measured) ground truth and estimated position and orientation trajectories of the tip. 
    Since the robot exhibited planar motions, there was only one coordinate ($\theta$) for the orientation and two coordinates ($x$ and $y$) for the position. 
    All the estimates exhibited close tracking of the ground truth after 0.12 seconds.
    The RMSEs after 0.12 seconds were $0.0069~\mathrm{rad}$ for $\theta$, $0.0026~\mathrm{m}$ for $x$, and $0.0064~\mathrm{m}$ for $y$.}
    \label{fig:tip pose estimation}
\end{figure}

\begin{figure}[t]
    \centering
    \includegraphics[width=0.98\columnwidth]{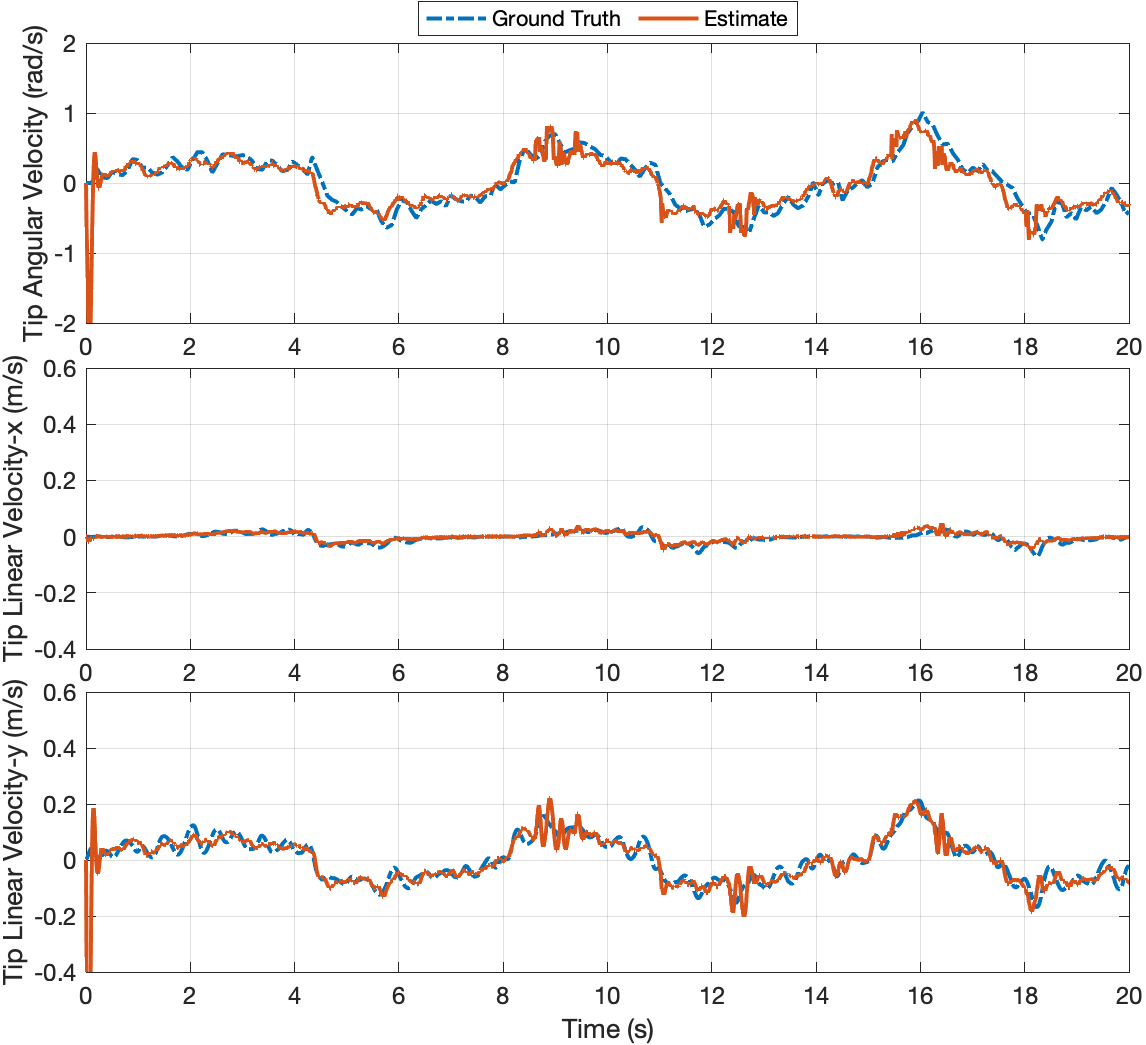}
    \caption{Comparison of the (measured) ground truth and estimated linear and angular velocity trajectories of the tip. 
    The estimated velocities reasonably tracked the ground truth after 0.2 seconds, although they exhibited more significant oscillations than the ground truth.
    The RMSEs after 0.2 seconds were $0.1345~\mathrm{rad/s}$ for the angular velocity, $0.0092~\mathrm{m/s}$ for the $x$ linear velocity, and $0.0270~\mathrm{m/s}$ for the $y$ linear velocity.}
    \label{fig:tip velocity estimation}
\end{figure}

\begin{figure}[t]
    \centering
    \includegraphics[width=0.98\columnwidth]{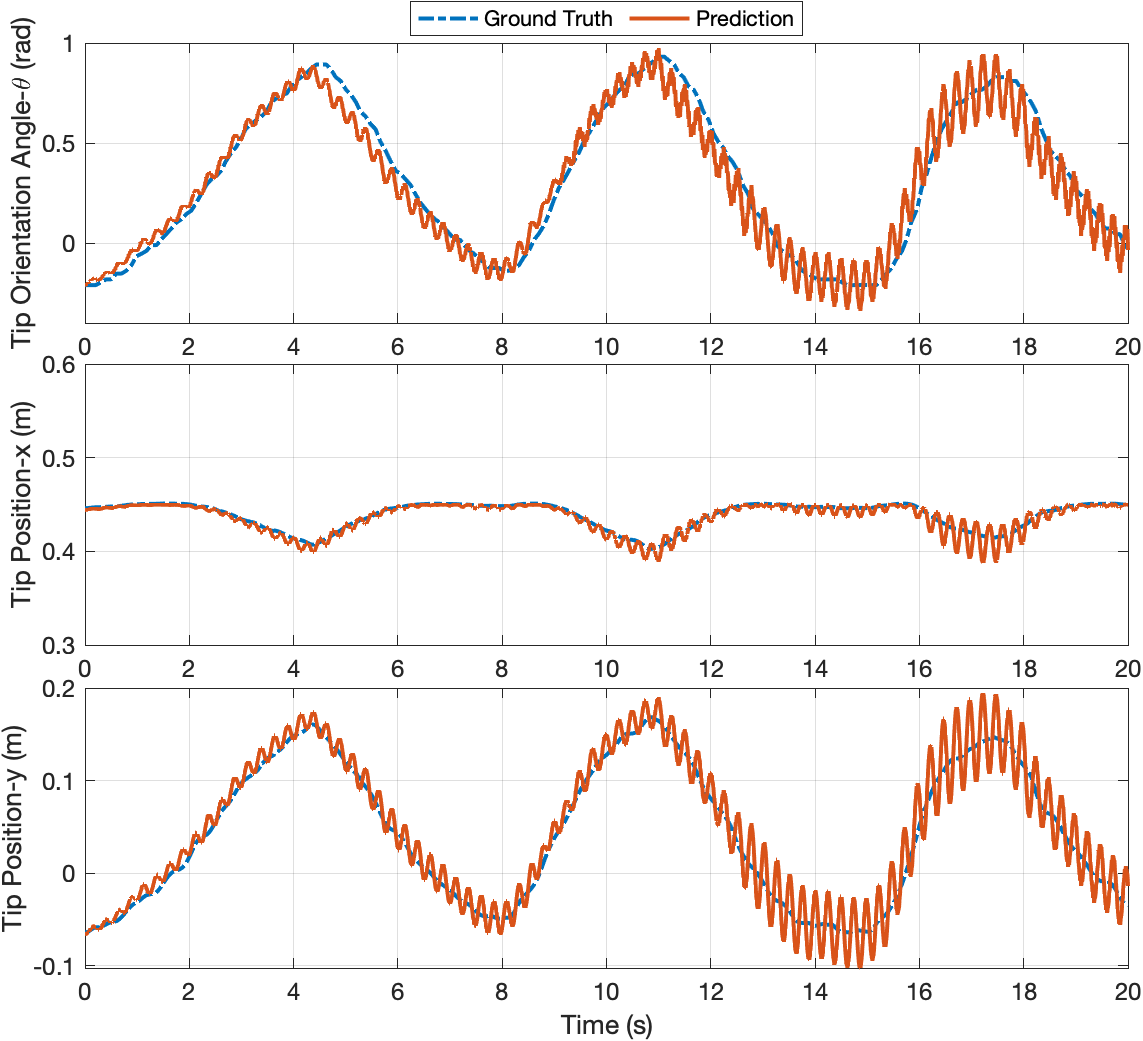}
    \caption{Comparison of the (measured) ground truth and predicted position and orientation trajectories of the tip. 
    Even with the exact initial estimate, pure model predictions exhibited increasingly accumulated deviations from the ground truth and more significant oscillations than the ground truth.}
    \label{fig:tip pose prediction}
\end{figure}

\subsection{Experiments and Algorithm Implementation}

A weight of 50 g, which was assumed to be known, was attached to the tip.
This simulated the scenario where a continuum robot is used to pick up an object.
The robot was initially at rest, which was a bending configuration due to its own weight.
The body frame of the base disk was defined by the red axes attached to it, which was also used as the spatial frame; see Figure~\ref{fig:Setup Diagram}.
The body frames attached to other cross-sections were similarly defined.
We held the force sensor and manually pulled the tendon for 20 seconds to generate motions in the $xy$ plane.
All data (force sensor readings, LED marker positions, and IMU measurements) were recorded on a PC for offline validation.
Both the force sensor and the IMUs had a sampling rate of 100 Hz.
The PhaseSpace motion capture system recorded the markers' positions (including the tip) at 960 Hz, which were then downsampled to 100 Hz.
The tip linear velocity was determined in the spatial frame using the down-sampled position recordings. It was then transformed into the body frame using the recorded rotation matrix from the IMU. To reduce noise, a moving average with a window size of 5 points was applied.

To implement the boundary observer (Eqns. \ref{eq:boundary observer 1}-\ref{eq:intermediate variable 2}), we needed to know the model parameters $J$ and $K$, which we calculated based on Table \ref{tab:robot parameter}.
We simply adopted all nominal values except for the density, which was calibrated from $2700~\mathrm{kg/m^3}$ to $20321~\mathrm{kg/m^3}$ to include the weights of the nine disks closest to the tip.
While one could (and usually should) further calibrate other nominal parameters using experiments, as suggested in \cite{rucker2011statics}, we skipped the fine calibration because our estimation algorithm is expected to be robust to this model uncertainty.
Denote the angular and linear inertia (stiffness) matrices by $J_1$ and $J_2$ ($K_1$ and $K_2$).
They were calculated according to
\begin{align*}
    J_1 & =\mathrm{diag}(2,1,1)\rho\pi r^4/4, \quad J_2  =I_{3\times3}\rho\pi r^2 , \\
    K_1 & =\mathrm{diag}(2G,E,E)\pi r^4/4, \quad K_2  =\mathrm{diag}(E,G,G)\pi r^2.
\end{align*}
We have $J=\mathrm{diag}(J_1,J_2)$ and $K=\mathrm{diag}(K_1,K_2)$.

We used SoRoSim, a MATLAB-based solver of Cosserat rod models, to implement our estimation algorithm \cite{mathew2022sorosim}.
We selected ``ode15s'' in SoRoSim for discrete temporal integration.
Therefore, the update was not performed at a fixed rate, and whenever external data (like force readings or tip measurements) were needed, we used the most recent data.
The gains were set to $\Gamma_{\text{P}}=0.05I_{6\times6}$ and $\Gamma_{\text{D}}=0.05I_{6\times6}$.
The estimation algorithm was initialized with a deviated configuration purposely to validate its convergence property.

To highlight the importance of our boundary correction technique, we also investigated the case where no boundary correction was used.
This was implemented by setting the observer gains to $\Gamma_{\text{P}}=\Gamma_{\text{D}}=0$.
In this case, the algorithm was initialized with the exact initial condition, which was obtained by computing the equilibrium configuration of the robot under its own weight and the tip load.

\subsection{Results}
    
We plotted the ground truth positions of the markers and the estimated continuum position of the backbone by our algorithm for four time instants in Figure \ref{fig:estimation}.
The corresponding snapshots of the experiment were also included.
The position estimates showed a good match with the ground truth positions even though the estimation algorithm was incorrectly initialized.
Note that our estimation algorithm computes not only positions, but also orientations, strains, and velocities.
We compared the estimated and measured (treated as the ground truth) trajectories of position, orientation, linear velocity, and angular velocity of the tip, because it was the location with the largest estimation error among all LED locations.
These trajectories are plotted in Figures~\ref{fig:tip pose estimation} and \ref{fig:tip velocity estimation}.
In Figure~\ref{fig:tip pose estimation}, even with a purposely deviated initial estimate, the position and orientation estimation errors decreased to $10\%$ of the initial errors in 0.12 seconds.
After this 0.12-second transient phase, the estimated positions and orientation angles exhibited close tracking of their ground truth.
The root-mean-square errors (RMSEs) of the estimated trajectories after the transient phase were calculated to be $0.0069~\mathrm{rad}$ for the orientation angle $\theta$, $0.0026~\mathrm{m}$ for the position coordinate $x$, and $0.0064~\mathrm{m}$ for the position coordinate $y$.
The estimates occasionally exhibited large deviations from the ground truth, especially when the robot was under large-angle deformation.
This was possibly because when the deformation was large, the assumed linear constitutive law \eqref{eq:linear constitutive law} became invalid and introduced more modeling errors. 
Nevertheless, our estimation algorithm still produced reasonable estimates of the true states, which validated that our algorithm is robust to modeling uncertainty.
In Figure~\ref{fig:tip velocity estimation}, the velocity estimates also converged to a small neighborhood of the ground truth in 0.2 s and then exhibited reasonable tracking of the ground truth.
The RMSEs of the estimates after the transient phase were calculated to be $0.1345~\mathrm{rad/s}$ for the angular velocity, $0.0092~\mathrm{m/s}$ for the $x$ linear velocity, and $0.0270~\mathrm{m/s}$ for the $y$ linear velocity.
We also observed that the estimated velocities exhibited more significant oscillations than the ground truth velocities. 
Again, this was possibly caused by the assumed linear constitutive law \eqref{eq:linear constitutive law} in our estimation algorithm.
In reality, the robot necessarily had material damping and therefore exhibited smaller oscillations.
Nevertheless, the reasonable tracking of the velocity estimates validated the robustness of our algorithm to this unmodeled material damping.

We also plotted in Figure~\ref{fig:tip pose prediction} the pure model prediction of the tip position and orientation by the same Cosserat rod model when the boundary correction technique was disabled.
We observed that even with the exact initial estimate, pure model predictions produced increasingly accumulated deviations from the ground truth over time.
The accumulated errors could result from modeling errors and measurement errors.
They suggested that it is not a good idea to use a dynamic model for long-time prediction without incorporating any additional sensing.
The pure model predictions also exhibited more significant oscillations than the ground truth.
This was expected because oscillations are natural in many second-order mechanical models if no material damping is assumed.
The comparison of Figures~\ref{fig:tip pose estimation} and \ref{fig:tip pose prediction} validated that our boundary correction technique based on the tip measurements was critical to consistently steering the estimation errors toward convergence and compensating different sources of errors, such as modeling errors, measurement errors, and inaccurate constitutive laws.

\section{Conclusion}
\label{section:conclusion}

In this work, we reported a generalization of our previous state estimation algorithm for continuum robots.
This algorithm was able to recover all infinite-dimensional continuum robot states by only measuring the pose and velocity of the tip.
The effectiveness of the algorithm was validated offline using experimental data recorded from a tendon-driven continuum robot.
The experimental results showed that the state estimates by this algorithm quickly converged to the recorded ground truth states and exhibited close tracking of the robot's actual motion.
Our future work is to validate the algorithm for 3D motions and in real-time experiments.

\section*{Appendix: Notation in $SO(3)$ and $SE(3)$}
\label{section:notation}

Denote by $SO(3)$ the special orthogonal group (the rotation matrices group in $\mathbb{R}^{3\times3}$) and by $so(3)$ its associated Lie algebra (the vector space of matrix logarithms in $\mathbb{R}^{3\times3}$ of rotation matrices).
Denote by $SE(3)=SO(3)\times\mathbb{R}^3$ the special Euclidean group (the group of homogeneous transform matrices in $\mathbb{R}^{4\times4}$) and by $se(3)$ its associated Lie algebra, (the vector space of matrix logarithms in $\mathbb{R}^{4\times4}$ of homogeneous transform matrices).
A hat $\wedge$ in the superscript of a vector $\eta$ defines a matrix $\eta^\wedge$ whose definition depends on the dimension of $\eta$.
Specifically, if $\eta\in\mathbb{R}^3$, then $\eta^\wedge\in so(3)$ is such that $\eta^\wedge\xi=\eta\times\xi$ for any $\xi\in\mathbb{R}^3$ where $\times$ is the cross product.
In this case, $\eta^\wedge$ turns out to be a skew-symmetric matrix in $\mathbb{R}^{3\times3}$.
If $\eta=[w^T,v^T]^T\in\mathbb{R}^6$ with $w,v\in\mathbb{R}^3$, then $\eta^\wedge\in se(3)$ is defined by
\begin{align*}
    \eta^\wedge=
    \begin{bmatrix}
        w^\wedge & v \\
        0_{1\times3} & 0
    \end{bmatrix}\in\mathbb{R}^{4\times4}.
\end{align*}
Let the superscript $\vee$ be the inverse operator of $\wedge$, i.e., $(\eta^\wedge)^\vee=\eta$.
The adjoint operator $\ad$ of $\eta=[w^T~v^T]^T\in\mathbb{R}^6$ with $w,v\in\mathbb{R}^3$ is defined by
\begin{align*}
    \ad_\eta=
    \begin{bmatrix}
        w^\wedge & 0_{3\times3} \\
        v^\wedge & w^\wedge
    \end{bmatrix}\in\mathbb{R}^{6\times6}.
\end{align*}

\section*{Acknowledgement}
Tongjia Zheng expresses gratitude to Federico Renda and Anup Teejo Mathew for their help in optimizing SoRoSim.

\bibliographystyle{IEEEtran}
\bibliography{References}

\end{document}